\documentclass[nohyperref]{article}

\AtBeginDocument{%
  \providecommand\BibTeX{{%
    \normalfont B\kern-0.5em{\scshape i\kern-0.25em b}\kern-0.8em\TeX}}}

\usepackage{url}

\usepackage[breaklinks]{hyperref}

\usepackage[utf8]{inputenc}
\usepackage{graphicx}
\usepackage[labelfont=bf, font=footnotesize]{caption}
\usepackage{subfigure}
\usepackage{microtype}
\usepackage{booktabs}

\usepackage{amsthm}
\usepackage{amsmath}
\usepackage{amssymb}
\usepackage{float}
\usepackage{xcolor}
\usepackage{nameref}
\usepackage{appendix}
\usepackage{titlepic}
\usepackage{mathtools}

\usepackage{etoolbox,lipsum}
\usepackage{multirow}

\usepackage[accepted]{dynamic_workshop/icml2022}

\usepackage[capitalize,noabbrev]{cleveref}

\theoremstyle{definition}
\newtheorem{condition}{Bias Condition}
\newtheorem{extension}{Bias Extension}
\newcommand{\R}{\mathbb{R}}

\theoremstyle{plain}

\theoremstyle{definition}

\theoremstyle{remark}

\usepackage[textsize=tiny]{todonotes}

\icmltitlerunning{Prisoners of Their Own Devices: How Models Induce Data Bias in Performative Prediction}

\begin{document}

\twocolumn[
\icmltitle{Prisoners of Their Own Devices: \\ How Models Induce Data Bias in Performative Prediction
           }
           


\begin{icmlauthorlist}
\icmlauthor{José Pombal}{Feedzai,IST,IT}
\icmlauthor{Pedro Saleiro}{Feedzai}
\icmlauthor{Mário A.T. Figueiredo}{IST,IT}
\icmlauthor{Pedro Bizarro}{Feedzai}
\end{icmlauthorlist}

\icmlaffiliation{Feedzai}{Feedzai}
\icmlaffiliation{IST}{Instituto Superior Técnico, Universidade de Lisboa}
\icmlaffiliation{IT}{Instituto de Telecomunicações}

\icmlcorrespondingauthor{José Pombal}{jose.pombal@feedzai.com}

\icmlkeywords{Machine Learning, ICML}

\vskip 0.3in
]



\printAffiliationsAndNotice{}

\begin{abstract}
    \noindent
    The unparalleled ability of machine learning algorithms to learn patterns from data also enables them to incorporate biases embedded within.
    A biased model can then make decisions that disproportionately harm certain groups in society.
    Much work has been devoted to measuring unfairness in static ML environments, but not in dynamic, performative prediction ones, in which most real-world use cases operate.
    In the latter, the predictive model itself plays a pivotal role in shaping the distribution of the data.
    However, little attention has been heeded to relating unfairness to these interactions.
    Thus, to further the understanding of unfairness in these settings, we propose a taxonomy to characterize bias in the data, and study cases where it is shaped by model behaviour.
    Using a real-world account opening fraud detection case study as an example, we study the dangers to both performance and fairness of two typical biases in performative prediction: distribution shifts, and the problem of selective labels.
\end{abstract}

\section{Introduction}\label{sec:intro}
With the increasing prominence of machine learning in high-stakes decision-making processes, its potential to exacerbate existing social inequities has been a reason of growing concern~\citep{exacerbate-inequities-howard2018ugly,exacerbate-inequities-kirchner2016machine,exacerbate-inequities-o2016weapons}. 
%
%
The goal of building systems that incorporate these concerns has given rise to the field of fair ML, which has grown rapidly in recent years~\citep{mehrabi2019survey}.

Fair ML research has focused primarily on devising ways to measure unfairness~\citep{measure-barocas} and to mitigate it in static algorithmic predictive tasks~\citep{mehrabi2019survey,exacerbate-inequities-caton2020fairness}.
However, the vast majority of real-world use cases operate in dynamic environments, which 
feature complex, and unpredictable feedback loops 
that may exacerbate existing biases in the data and models.
%
In such environments, model behaviour itself shapes the distribution of the data, so a deep understanding of data bias and its interaction with the ML model is required to uncover the causes of unfairness. 
That said, accounting for some notable exceptions~\citep{labels-fogliato2020fairness, datasource-noisylabelswang, similar-blanzeisky2021algorithmic, harvard-workshop-relating, cmu-bias-stress} 
little attention has been heeded to relating unfairness to concrete bias patterns in the data.
%


To this end, we propose a domain-agnostic taxonomy to characterize data bias between a protected attribute, other features, and the target variable.
It may be applied in dynamic environments, where bias during training can lie in stark contrast with that found in production.
In particular, we use the taxonomy to model performative prediction settings, where data bias is induced by the predictive model itself.
As a running example, we take an account opening fraud detection case study, which features two typical performative prediction bias phenomena:
first, distribution shifts from fraudsters adapting to escape the fraud detection system;
second, noisy labels arising from the selective labels problem, where the AI system determines the observed labels.
We show how both issues, if left unaddressed, have detrimental, unpredictable, and sometimes unidentifiable consequences on fairness and performance.
%

\section{Background \& Related Work}\label{sec:rw}

\citet{performative-prediction} define predictions as performative if they ``influence the outcome they aim to predict''.
This influence usually reflects itself in distribution shifts over time, which, if left unaddressed, lead to degradation in predictive performance.
In a lending scenario, \citet{performative-prediction-jpmorgan} study the effects on fairness metrics of a classifier whose prediction changes the lending approval probability for a protected group.
\citet{strategic-fairness} point out the negative impact that adaptive agents have on the effectiveness of classifiers trained to be fair.
Conversely, our work focuses on group-wise feature distribution shifts due to fraudsters adapting to the model over time --- a \textit{strategic classification}~\citep{strategic-classification} setting, which is a subset of performative prediction. 
%

The selective label problem arises when the system under analysis determines the sample of observed labels~\citep{fairmlbook}.
For example, in fraud detection, rejecting an account opening based on the belief that it is fraudulent, implies that we will never observe its true label (the account never materializes).
Here, the system fully determines the outcome of that observation, making it performative. 
Assuming the labels of these instances to be truthful causes a bias in the evaluation of performance and fairness of classification tasks, since most metrics --- and particularly those one which we focus here (e.g.: FPR) --- rely on accurate ground-truth labels.
There is some work discussing the detrimental effects of noisy labels on algorithmic fairness~\citep{labels-fogliato2020fairness, labels-jiang2020identifying, labels-lamy2019noise, datasource-noisylabelswang,datasource-noisylabelsfair}.
%
The work of \citet{label-shift} is particularly pertinent, as it discusses the impact of label noise and shift on the reliability of fairness-aware algorithms.
Our paper extends that work to a setting where label bias is induced by the classifier itself.



\section{Bias Taxonomy}\label{sec:definitions}
Throughout this work, we refer to the features of a dataset as $X$, the class label as $Y$, and the protected attribute as $Z$. 
%
%
The following definitions use the inequality sign ($\neq$) to mean a statistically significant difference.
%
%

 
Despite a multitude of definitions, there is still little consensus on how to characterize data bias 
\citep{mehrabi2019survey}.  
In this paper, we propose a broad definition: there is bias in the data with respect to the protected attribute, whenever the random variables $Y$ and $X$ are sufficiently statistically dependent on $Z$.
This does not mean a classifier trained on such data would be unfair, but rather that there is potential for it to be.
In Section~\ref{sec:results-dynamic}, we will explore how these biases may be induced by model behaviour over time.

\textbf{Base Bias Condition}
    \begin{equation}\label{eq:cond-broad-bias}
        P[X, Y] \neq P[X, Y | Z].
    \end{equation}
To satisfy this, $Z$ must be statistically related to either $X$, $Y$, or both. 
The following biases imply this condition.
%
%

%

\textbf{Group-wise Class-conditional Distribution Bias}
    \begin{equation}\label{eq:cond-condist}
        P[X|Y] \neq P[X | Y, Z].
    \end{equation}
    Consider an example in account opening fraud in online banking.
    Assume that the fraud detection algorithm receives a feature which represents how likely the email-address is to be fake (X) and the client's reported age (Z) as inputs.
    In account opening fraud, fraudsters tend to impersonate older people, as these have a larger line of credit to max out, but use fake e-mail addresses to create accounts.
    %
    %
    Therefore, the e-mail address feature will be better to identify fraud instances for reportedly older people, potentially generating a disparity in group-wise error-rates, even if age groups have an equal likelihood of committing fraud in general.

\begin{figure}[tb]
    \centering
    \includegraphics[width=1\linewidth]{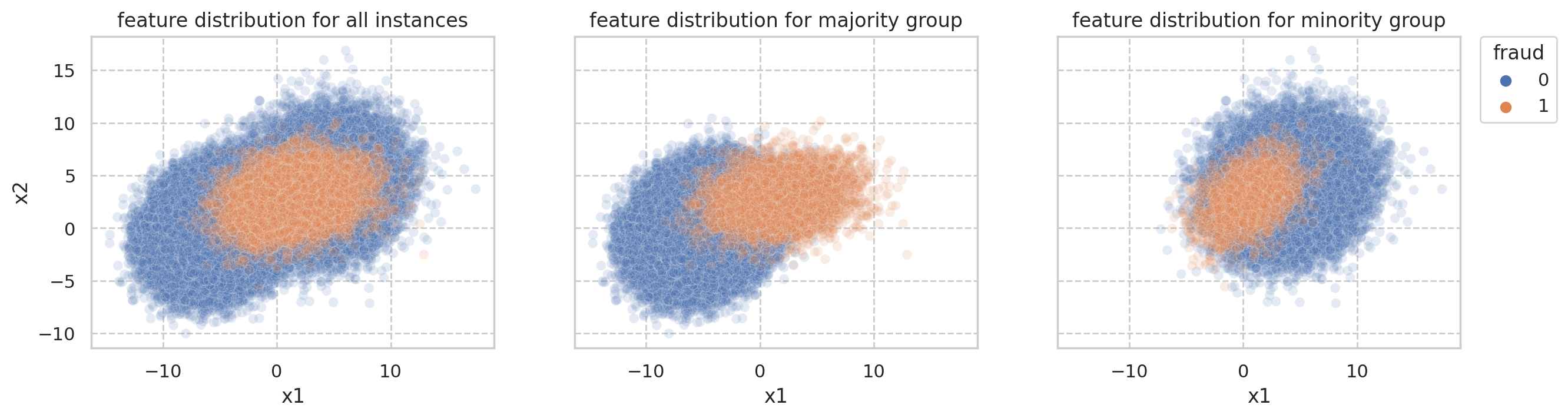}
    \caption{
    Group-wise Class-conditional Distribution Bias. There is clear class separability for the majority group (middle), \textit{i.e.}, we can distinguish the fraud label using the two features.
    At the same time, there is virtually no separability for the minority group (right), as positive and negative samples overlap on this feature space.
    However, this is not discernible when looking at the marginal distribution for Y, $x_1$, and $x_2$ (left).
    }\label{fig:separability}
\end{figure}

    %
\textbf{Noisy Labels Bias}
    \begin{equation}\label{eq:cond-noisy-labels}
        P^*\left[Y \middle| X, Z\right] \neq P\left[Y \middle| X, Z\right],
    \end{equation}
    where $P^*$ is the observed distribution and $P$ is the true distribution.
    That is, some observations belonging to a protected group have been incorrectly labeled.
    %
    It is common for one protected group to suffer more from this ailment, if the labeling process is somehow biased. 
    For example, women and lower-income individuals tend to receive less accurate cancer diagnoses than men, due to sampling differences in medical trials~\citep{noise_medical_trials}.
    In fraud detection, label bias may arise due to the selective label problem.
    %

\textbf{Dynamic Bias}

    Let $ \textbf{BC}_i $ be a set of bias conditions \textbf{BC} on a data distribution $i$. Then, under dynamic bias, biases may change over time such that,
    \begin{equation}
        \boldsymbol{BC}_{train} \neq \boldsymbol{BC}_{deployment} .
    \end{equation}
    This bias is the \textit{bread and butter} of dynamic environments, as one of the main challenges in these domains is adapting to the disparities between training and deployment data.
    Indeed, distribution shifts may greatly affect model performance and fairness.
    In fraud detection, this can be particularly important, if we consider that fraudsters are constantly adapting to the  model to avoid being caught.
    As such, a trend of fraud learned during training can easily become obsolete after deployment.

\section{Case Study}\label{sec:results-dynamic}
\subsection{Dataset and Methodology}\label{subsec:dataset}
In this work, we use a real-world large-scale case study of account-opening fraud (AOF).
%
Each row in the dataset corresponds to an application for opening a bank account, submitted via the online portal of large retail bank.
Data was collected over an 8-month period, and contains over 500K rows.
The first 6 months are used for training and the remaining 2 months are used for testing, mimicking the procedure of a real-world production environment (we change this in Section~\ref{subsec:UC2}).
As a dynamic real-world environment, some distribution drift is expected along time, both from naturally-occurring shifts in the behavior of legitimate customers, as well as shifts in fraudsters' illicit behavior as they learn to better fool the production model.

Fraud rate (positive label prevalence) is about $ 1\% $ in both sets.
This means that a naïve classifier that labels all observations as \textit{not fraud} achieves a test set accuracy of almost $99\%$.
Such large class imbalance entails certain additional challenges for learning~\citep{imbalanced-learning} and calls for a specific evaluation framework.
As such, performance will be measured as true positive rate (TPR) at a threshold of 5\% false positive rate (FPR).
TPR measures the percentage of detected fraud, and the FPR is limited at 5\% as usually required by banks to avoid customer attrition (each FP is a legitimate application flagged as fraudulent, which can cause customers to want to change banks).

We will measure fairness as the ratio between group-wise FPR, also known as \textit{predictive equality}~\citep{Corbett-Davies2017}, which measures whether the probability of a legitimate person being flagged as fraudulent depends on the group they belong to.
%
%
This fairness measure is by no means perfect, or enough to ensure fairness in many senses, but given our \textit{punitive} setting, it is considered appropriate~\citep{Corbett-Davies2017, Saleiro2018}.

\subsection{Scenario 1: Adaptive Fraudsters}~\label{subsec:UC1}
Fraud detection is a typical case of performative prediction for two reasons.
First, the system determines the outcome of instances it flags as fraud: they are blocked, and so fraud never materializes.
Second, fraudsters (the target of the predictor) actively adapt to evade the fraud detection system.
This response emerges in the form of distribution shifts, where certain useful patterns to detect fraud in training become obsolete in production (post-deployment).
%
%
Extending these concerns to fairness is straightforward, if we assume that fraudsters may leverage certain sensitive attributes in tandem with other features to escape detection.
Indeed, one can use our proposed data bias taxonomy to model this as a combination of Group-wise Class-conditional Distribution Bias and Dynamic bias.
%

To illustrate this, given a synthetic binary protected attribute, we add two synthetic features $x_1$ and $x_2$ to our dataset, which made it easier to detect fraud for one protected group during the past, and is reflected in the current dataset.
%
However, at test time, this groups' fraudsters adapt to the system, rendering these features useless for predicting fraud.
Group sizes and fraud rates are made equal, such that there is no other bias in the data.
We compare the performance and fairness of 50 LighGBM models under this setting (Adaptation) with two other cases: one where fraudsters did not change their behaviour (Performance Ideal), and one where the additional features did not exist (Unbiased Baseline).

\begin{figure}[tb]
    \centering
    \includegraphics[width=\linewidth]{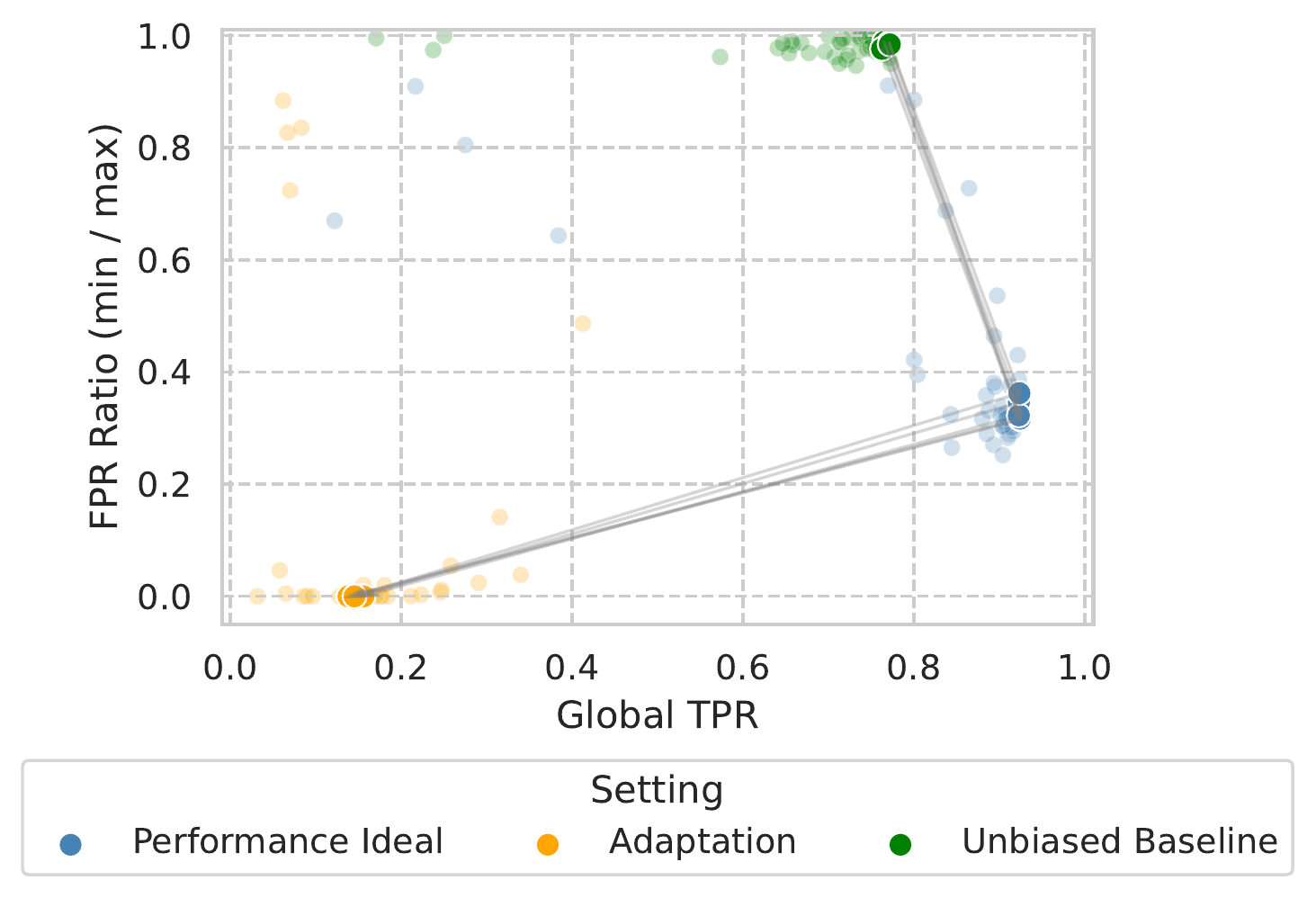}
    \caption{
    The opaque points in blue are the top 5 performing models on the setting in which the practitioner believes they are operating (one of these models would be chosen for production).
    The opaque points in yellow show the same models on the performative prediction setting on which the practitioner is actually operating, where fraudsters adapted their behavior.
    The opaque points in green are the models trained on a baseline setting, where the protected attribute is independent of X and Y.
    Arrows connect these model configurations, showing how the ones selected for production under `Performance Ideal' are not the best in `Adaptation'.
    }
    \label{fig:use_case_1}
\end{figure}

Figure~\ref{fig:use_case_1} shows how fraudsters adapting to the system in production had a harmful impact on both performance and fairness\footnote{the extent of this degradation could have been smaller or larger, depending on the nature of the distribution shift.} of the top performing models.
The former was to be expected, but the latter is somewhat surprising.
Given that the additional features became uninformative in testing, it would have been desirable that the models converged to the performance and fairness equilibrium of the Unbiased Baseline, which did not have $x_1$ and $x_2$.
However, models were ``lazy'', and did not learn some of the useful fraud patterns in the already-present features.
Instead, they focused mostly on $x_1$ and $x_2$, missing out on a chance to increase both fairness and performance. 
Notice how the best models on ``Performance Ideal'' were not the best, or the fairest, after the fraudsters' adaptation.
This highlights the importance of using ML methods that are robust to distribution shifts, especially in performative prediction environments.
One such method would have been able to improve both fairness and performance.

\subsection{Scenario 2: Selective Noisy Labels}\label{subsec:UC2}

Fraud detection suffers from an issue of selective labels, whereby the practitioner never gets to verify the true label of instances that are classified as fraud (predicted positives).
For example, if we block the opening of an account, we can never confirm whether it would have been a fraudulent instance.
Still, it is common practice to re-use these predictions in the training of future models as positive label examples.
If one admits that a fraction of these observations are false positives, models will learn on noisy labels as time goes by.
Thus, this problem can be framed as an instance of Noisy Labels Bias across time.
Contrarily, the label of predicted negatives is eventually revealed, since these either materialize into fraud, or are in fact not fraud.

To assess the impact of this noise on fairness, we set up an experiment which mimics real-world fraud detection systems.
We split the dataset into four temporal folds, starting with 3 months for training, 1 month for validation and 1 for testing.
At each of four sliding-window iterations of training and evaluating models, we advance by 1 month, concatenating the previous validation set onto training, using the previous test set as validation, and moving on to a more recent test set.
%
Importantly, the positive labels used to train and validate subsequent models are all the false negatives, and predicted positives of past models.
This injects the type of label noise we mentioned above, with false positives being noisy label positives. 
%
%
%

\begin{figure}[tb]
    \centering
    \includegraphics[width=\linewidth]{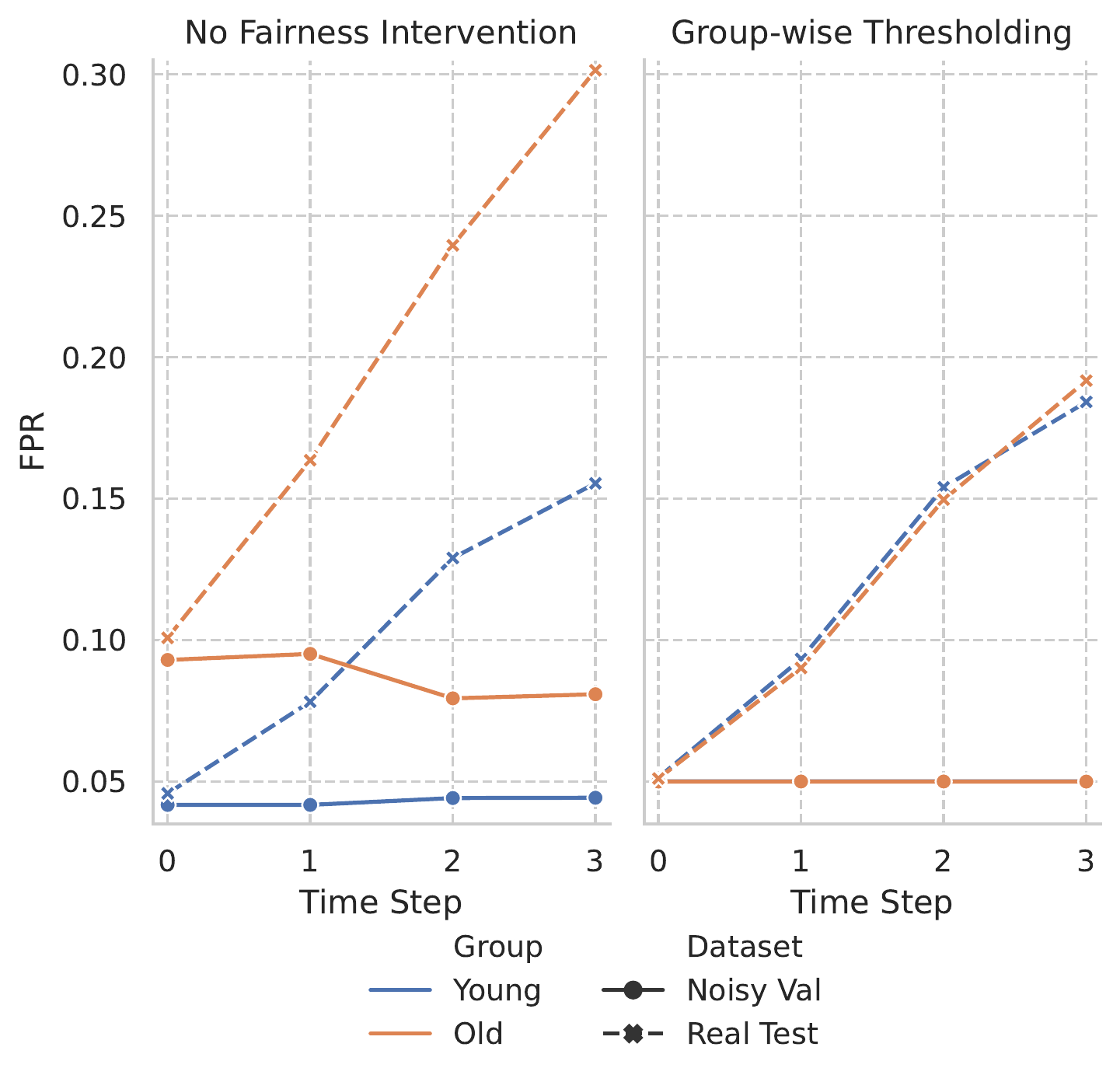}
    \caption{FPR over 4 time steps. Time Step 0 has no noise, but from then on some labels are noisy. The practitioner believes they are operating at 5\% FPR (Noisy Val), but in production (Real Test), the FPR calculated on the true labels is much higher. 
    }
    \label{fig:use_case_2}
\end{figure}

%
%
%
Two settings of the experiment are run.
In one, a global threshold to achieve 5\% FPR on the noisy validation set is used.
This is the standard for fraud detection, as used in Scenario 1.
In the other, we use group-wise thresholds --- a popular post-hoc fairness intervention~\citep{hardt2016equality} --- such that \textit{predictive equality} (equal group FPRs) is satisfied on the validation set.
The goal is to assess whether unfairness observed in the first setting can be mitigated, or if the selective label bias renders the intervention fruitless.
If the latter is the case, practitioners should tackle the selective label problem before trying to guarantee fairness.
At each iteration, the best performing LightGBM on the validation set over 50 trials of TPE hyperparameter optimization \citep{tpe} is evaluated.

Figure~\ref{fig:use_case_2} compares group-wise FPRs after thresholding model scores in the noisy validation set, versus the FPRs the same model obtains in production (test set), when evaluated on real labels.
Conditioned on the group, FPRs start off only slightly different due to natural distribution shifts between validation and test sets. 
However, they diverge as noise levels grow, increasing three-fold between the first and last iterations (30\% v.s. 8\% for the ''Old`` group, and 15\% v.s.  15\% for the ''Young`` group).
The gap between blue and orange lines also widens in production, meaning higher levels of unfairness (about which the practitioner is unaware).
%
%
Not only do these phenomena have harmful consequences on business, but the increased unfairness contributes to aggravate existing societal inequities.
Even after the fairness intervention, the rift in group-wise FPRs shows a tendency to widen as label noise accumulates.
%
%
Thus, mitigating the selective labels problem is of paramount importance in ensuring that systems are in fact fair in dynamic settings.
%
We also tried dropping older training observations, a common industry practice due to storage and computational limitations.
This seemed to attenuate the gap between perceived and real FPR, but to widen the disparity in group FPRs. 

\section{Conclusion}~\label{sec:conclusions}
We proposed a data bias taxonomy to characterize the causes of unfairness in dynamic environments, where models shape the data distribution.
In particular, we modelled two scenarios of bias in performative prediction: strategic classification, and selective noisy labels.
We showed how both issues, if left unaddressed, have detrimental, unpredictable, and sometimes unidentifiable consequences on fairness and performance.
We hope this work inspires future research on developing suitable fairness interventions for dynamic environments.

%

\clearpage
\bibliographystyle{dynamic_workshop/icml2022}
\bibliography{refs}

\end{document}